\documentclass{article}

\usepackage[final,nonatbib]{nips_2016}

\usepackage[utf8]{inputenc} 
\usepackage[T1]{fontenc}    
\usepackage{varioref}
\usepackage[colorlinks=true,linkcolor=red]{hyperref}
\usepackage{cleveref}
\usepackage{url}            
\usepackage{booktabs}       
\usepackage{amsfonts}       
\usepackage{nicefrac}       
\usepackage{microtype}      

\usepackage{amsmath}
\usepackage{amsthm}
\usepackage{amsfonts}
\usepackage{booktabs}       
\usepackage{todonotes}
\usepackage{subcaption}
\newcommand{\bvec}[1]{\mathbf{#1}}

\title{Automatic Discoveries of Physical and Semantic Concepts via Association Priors of Neuron Groups}

\author{Shuai Li\\
  Chinese University of Hong Kong\\
  \texttt{sali@ee.cuhk.edu.hk} \\
  \And
  Kui Jia\\
  South China University of Technology
  \\
  \texttt{kuijia@scut.edu.cn} \\
  \And
  Xiaogang Wang\\
  Chinese University of Hong Kong\\
  \texttt{xgwang@ee.cuhk.edu.hk} \\
}

\begin{document}

\maketitle 



\maketitle

\begin{abstract}
  The recent successful deep neural networks are largely trained in a
supervised manner. It {\it associates} complex patterns of input samples with
neurons in the last layer, which form representations of {\it concepts}. In
spite of their successes, the properties of complex patterns associated a
learned concept remain elusive. In this work, by analyzing how neurons are
associated with concepts in supervised networks,
we hypothesize that with proper priors to regulate learning, neural networks
can automatically associate neurons in the intermediate layers with concepts
that are aligned with real world concepts, when trained only with
labels that associate concepts with top level neurons, which is a plausible way
for unsupervised learning. We develop a prior to verify the hypothesis and
experimentally find the proposed prior help neural networks
automatically learn both basic physical concepts at the lower layers, e.g.,
rotation of filters, and highly semantic concepts at the higher layers, e.g.,
fine-grained categories of an entry-level category.
\end{abstract}

\section{Introduction}
\label{sec:introduction}

Deep neural networks \cite{LeCun2015} have recently brought remarkable
breakthroughs in high dimensional perception problems such as image
classification \cite{Krizhevsky2012} and object detection \cite{He2016}. Most
of these successes are achieved in the context of supervised learning where
tremendous amounts of labeled training samples are provided. Based on training
objectives that directly {\it associate} training samples with their semantic
labels, supervised learning learns neural networks that associate complex
patterns of input samples to hidden neurons in the last layer, which form
representations of {\it concepts}. These hidden neurons are subsequently fed to
a classifier to estimate the probability of semantic labels. In spite of their
successes, the properties of complex patterns associated a learned concept
remain elusive.

\begin{figure}[h]
  \centering
  \includegraphics[width=0.8\linewidth]{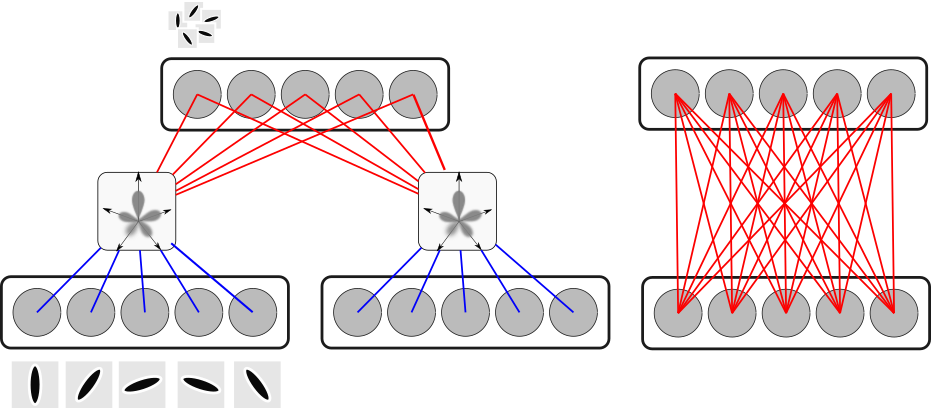}
  \caption{Illustration of Concept; of Neuron as representation of Concept; and
    of the proposed prior to regulate learning of Neural Network and its comparison
    with traditional neural networks. Concept is the cluster of images that
    contains variations of a prototype, e.g. the concept of an edge that consists of
    various orientations of an edge, shown at the lower left corner, and technically defined as the orbit
    generated by symmetry groups. Now we take
    orbit, concept and cluster as synonyms. Neurons in this layer are
    variations of concepts of neurons in the next layer. A neuron in
  the next layer is built through convolution
i.e. the red lines, and serves as the representation of a concept that is
associated with/invariant to
  variations of the concept (thanks to cross channel summation in convolution),
  e.g. as the concept shown above the top left neuron, a neuron in the next layer may be the concept of an edge, and will
  be active regardless of which orientation of edges is present.
  Illustrated on the left, we regulate the
  learning of neural networks with a prior that divides neurons into groups,
  and endorse the statistic prior on them that only one neuron/variation will be active,
  denoted as the probability density cloud in the square frame. Each arrow in
  the frame denotes a variation of a concept, and the probability density
  denotes how probably the current input sample contains that variation. Compared with
  traditional neural network, shown on the right, interaction among neurons in
  the same layer is present, denoted as blue lines. In modeling their interaction, we hope we can
  group variations of the same concept into the same group, e.g. all neurons in
  the group represent edges with various orientations. These variations
  are themselves more fine-grained concepts, which we consider may be aligned with
  real world concepts even if we do not train networks with labels of these
concepts.
  }
\label{fig:illustration}
\end{figure}

In this work, by analyzing how neurons are associated with concepts in
supervised networks, we hypothesize if with proper priors to regulate the
learning of neural networks, networks can associate filters with concepts
without supervised labels, which is a plausible approach to unsupervised
learning. To reach this hypothesis, we analyze how neural networks associate
neurons with concepts, as illustrated in \Cref{fig:illustration}. We define Concept formally in \Cref{sec:concept}, which
allows us to analyze how a neuron is built as a representation of a concept in
neural networks in
\Cref{sec:neural_concept}. The analysis implies that supervised labels can be
understood as a prior that associates neurons with concepts sample by sample, and leads to
our hypothesis.  To initially investigate the hypothesis, we observe that a
rich hierarchical structure exists in concepts --- concepts are built on top of
variations of concepts, which in turn is a concept themselves, e.g. the concept
of aquatic animals is built on top of fishes, seals etc, and fishes is a
variation of aquatic animals. We further hypothesize that
with proper priors to regulate learning, neural networks can automatically
associate neurons in the intermediate layers with variations of a concept that
are aligned with real world concepts, when trained only with labels that associate
concepts with top level neurons.  To verify, we develop a prior to regulate the
learning of networks in \Cref{sec:hypothesis}. It divides neurons into groups,
and aim to group variations of the same concept in the same group via
promoting competition among them. Then by only providing labels at the
top granularity of a concept, we test whether concepts of lower granularity can
also be learned without direct labels. Related works are discussed in \Cref{sec:related_works}.



In \Cref{sec:experiments}, we report experimental results. We train supervised neural networks with the proposed prior, and investigate
learned filters and neurons to see whether they can be associated with known
concepts. By applying the prior to a neural network, we find through filter
visualization filters in the same group of the first layer tend to be physical
variations of each other, the relationship among which probably forms a symmetry
group. In a problem where classes of training samples
have fine-grained class structures (i.e. a class has its underlying
sub-classes), with proposed prior, we find neural networks are able to
associate a filter with a sub-class variation within a super-class
automatically (in this problem case, we also call the classes with training
labels as common or super classes). More specifically, the prior is applied on
the penultimate layer. The last layer is a degenerated linear filter (max
pooling) layer that computes representations of super-class concepts. Training
labels associate neurons there with concepts of super-classes. After
training, we find neurons in the penultimate layer, which are trained with no labels,
discover clusters of samples that are particularly relevant to sub-class
variations in the super-class. When using discovered neurons to classify test
samples, we obtain classification performance that is far beyond the accuracy
of chance and approaches that of fully supervised learning. In contrast, no
clusters relevant to the sub-classes emerge from internal neurons of neural
networks without using our proposed association prior.



\section{Concept}
\label{sec:concept}

In this section, we define {\bf Concept} using symmetry group theory.

{\bf Concept as variations of a prototype}. First, we give an informal
definition that tries to capture the intuition behind the formal definition
given later. We define a cluster of images that consist of a prototype $t$ and
its variations $t', \forall t' \in T$ as the concept related to that prototype,
where $T$ is the set of all variations. To ease
discussion following, we call $t'$ a variation of a concept. It models the
behavior that humans distinguish or learn a novel concept by remembering
typical properties of a prototype of such concept, and are able to generalize
to its variations, e.g. an apple and its various viewpoints.

{\bf Symmetry group.} To give a formal but limited definition to $t, t'$, we
introduce Symmetry Group Theory. A symmetry of an object is a transformation
that leaves certain properties of that object intact \cite{Miller1972}. If we
have two symmetry transformation $g$ and $h$ and we compose them, the result
$gh$ is also a symmetry transformation. A symmetry group is a set with such a
composition operator on its elements. The composition needs to satisfy closure
and associativity. Identity element and inverse of any member also need to
exist.

{\bf Concept as orbits of symmetry group}.
 With the machinery of Symmetry Group, we
define a concept as the orbit of a prototype generated by certain symmetry
groups. For a symmetry group $G$, the orbit is a cluster of images consists of
the prototype $t$, and its variations $gt, \forall g \in G$, where $g$ is the
operation that varies/transform $t$ in group $G$. The definition characterizes the relationship
between $t$ and $t'$. For strict subsets of variations that
are induced by known symmetry groups, e.g. rotation, of a prototype $t$, such
definition is strictly accurate.
In this case,
we have a clear definition of $g, G, t$, e.g. a rotation translation group can be
represented through matrix,
$
  g(r, u, v) =
\begin{bmatrix}
  \cos(r\pi/2) & -\sin(r\pi/2) & u\\
  \sin(r\pi/2) & \cos(r\pi/2) & v\\
  0 & 0 & 1
\end{bmatrix}
$, $G = \{g(r, u, v) | r, u, v \in Z^{3}\}$, and $t$ is a function $t: Z^{3}
\rightarrow R^{|Z^{3}|}$ defined on $Z^{3}$. $Z^{3}$ is a tuple composed of
three integers. From then on, we will treat orbit and concept as synonyms.


{\bf Limitation of symmetry groups definition.} However, it is unlikely that
more complex transformation has an explicit parametric form as that of a small
affine group, e.g. SO rotation group given before, considering that they need
more than just physical rules. As an example, the face viewpoint transformation
actually needs further information, e.g. the other side of the face, which has
to be learned. Probably new mathematical objects need to be created to
characterize such behaviors, starting from symmetry group theory.  Thus, the
above formal definition is limited, but it serves as concept reuse for the
coming discussion.

\section{Neural network associates neurons with concepts}
\label{sec:neural_concept}

In this section, based on the concept defined, we explain the hypothesis how
neural networks associate neurons with concepts.

\subsection{Neural network}
\label{sec:group-restr-boltzm}

Formally, a convolutional neural network is a composition of layers of
computational units which defines an operator

\begin{displaymath}
  T: x(u) \rightarrow x_L(u, c)
\end{displaymath}

where $x(u)$ is the input signal, such as an image $I \in
L^{2}(\mathbb{R}^{2})$, $x_L(u, c)$ is the last layer that would be fed to a
classifier; $u$ is the translation index, and $c$ is the channel index. $x_L(u,
c)$ is computed from $x(u)$ through composing representations layer by
layer. Representations $x_l(u, c)$ at layer $l$ is obtained from representations
$x_{l-1}(u, c)$ at layer $l-1$ by applying a linear operator and an activation
function:

\begin{displaymath}
  x_{l} = \eta(W_l x_{l-1})
\end{displaymath}



\subsection{Neurons are representations of concepts}
\label{sec:group-assoc-prior-1}

Theoretical and experimental works suggests neural networks encode complex
transformation. \cite{Mallat2016} implies that instead of learning transformations $g$ for complex
variations, neural networks directly learn transformed results --- orbit
elements $gt$. The orbits are generated from
discrete samplings of continuous lie groups at intervals such that the values
of continuous transformation can be interpolated in between --- lie groups are
continuous symmetry groups whose elements form a smooth differentiable manifold
--- though the properties of those groups remain largely unknown.
Experimentally, the community has observed and hypothesized that the feature
space of neural networks lies in certain manifolds/lie groups \cite{Brahma2015}
\cite{Gardner2015}. As examples, point can traverse in the feature space to
move from an image of a person from a young age to a senior age
\cite{Gardner2015}, which demonstrates filters somehow encode such age
transformation.

Inspired by above works, we hypothesize a neuron in $x_l$ serves as a
representation of a concept, where in a neural network, a cluster of samples of
various variations of a concept are mapped to when doing forward propagation.
The representation is built recursively as described in the following.



The cross-channel summation done when convoluting $x_{l-1}$ with $W_l$ is
possibly a general way to selectively filter elements of orbits
\cite{Mallat2016}, or in another word, reuse concepts, and build new representations of concepts from them. We
illustrate the effect of convolution with a degenerated linear filter --- max
pooling. A neuron $x_{l+1, G}$ after convoluting/linear filtering are

\begin{displaymath}
  x_{l+1, G} = \max_{\forall g_i \in G} \{ \langle x_{l-1}, g_i t \rangle \}
\end{displaymath}

where $gt$ denotes filters in layer $l$, linear filtering is written as inner
product, the filter in layer $l+1$ that corresponds to the neuron $x_{l+1, G}$,
whose functional form is $\max$ operator, has coefficients $1$ in the dimension
where $x_l$ has the large activation value, and $0$ otherwise. $\langle
x_{l-1}, g_i t \rangle$ checks whether the input has orbit element $g_it$, and
the $\max$ operator computes a representation of the orbit/concept that is
invariant to $G$.

$x_{l-1}, x_l$ are also built similarly and is invariant to smaller symmetry groups.
With a compositional structure, distributed regions in the input space can be
routed to the same neuron in a feature map $x_{l}(u, c)$. When $\eta$ is a
piecewise linear activation function, \cite{Montufar2014} theoretically
analyzes the expressive power of a hidden neuron in feedforward neural
networks. Consequently, each neuron of $x_l$ is capable of being invariant to a large $G$ built
hierarhically, and associates with a cluster of samples of all kinds of
variations of a concept.

\section{The proposed prior}
\label{sec:hypothesis}

According to the hypothesis, in supervised neural networks, each neuron in the
last layer of the network computes a representation of a concept,
e.g. Cat. Supervised labels can be understood as a prior that functions sample
by sample to associate a particular neuron with a cluster of variations of cat
images. An important question to ask is, is it possible to deduce more general principles
underlying the sample-wise priors, and associate neurons with concepts in an
unsupervised manner?

To initially explore the question, we observe that a rich hierarchical structure
exists in concepts --- concepts are built on top of variations of concepts,
which in turn is a concept themselves, e.g. the concept of aquatic animals is
built on top of fishes, seals etc, and fishers is a concept itself. We hypothesize that with proper priors to
regulate learning, neural networks can automatically associate filters in
the intermediate layers with variations of a concept that are aligned with known
concepts, when trained only with labels that associate concepts with top level
neurons.

We develop a prior to fulfil such tasks in this section.
The intuition of the prior is that only when a concept is novel enough, a
network bothers to generate another neuron to represent that concept.
More formally, since orbit elements are generated by discrete sampling of a lie
group on the hypothesis to guarantee continuous interpolation, only when two
orbit elements are different with each other considerably, it needs another
sampling to differentiate one from another.
Given the radical variations between neurons, correspondence between
neural-network-learned concepts and concepts in the real world can be found,
which we experimentally verify later.


\subsection{Competitive Group-restricted Boltzmann Machine}
\label{sec:group-restr-boltzm}

To characterize the intuition, we divide neurons into groups, and aim to group
variations of the same concept in the same group via promoting competition
among orbit elements that are generated by the same lie group.

It works by encouraging concepts to associate with neurons that have larger
activation compared with other neurons, and be picky with orbit elements.
The rationale is during training,
all neurons are different sightly at the beginning by random initialization.
The output of linear filtering is the auto-correlation score between an input
patch (an image patch or feature map patch of the previous layer) and the
filter. The larger the output, the more characteristic that neuron is for the
concept of that sample, so that neuron should more associate with that
sample. To put it in another way, for the concept this particular sample belongs to, it
should be represented with such neuron.
On the other hand, one widely
acknowledged observation that one function of hidden neurons is to detect
salient patterns \cite{Zeiler2014}, and passes relevant signals to the next
layer. Thus, one potential relationship between $gt, \forall g \in G$ is to
characterize this functionality by giving out the most representative $gt$ in
the orbit, consequently reducing neuron activation variability, and sparsifying
activations, which may lead to better linear separability, better
disentanglement of factors of variations and more representative neurons
\cite{Glorot2011}\cite{Mallat2016}. For example, the neurons that are
associated with different viewpoints of a face does not fire at the same time
(as a side note, two viewpoints of the same face in different locations of an
image is another matter which is modeled by translation symmetry).
After training, the competition may
lead to filters that are considerably different orbit elements, and for each
forward propagation, only the most salient element is passed on.
In
addition, competition is a canonical computation \cite{Carandini2012} in the
brain.




Let the joint probability of a set of hidden units $\bvec{h}$ and a set of
visible units $\bvec{v}$ of a Boltzmann machine be

\begin{displaymath}
  p(\bvec{h, v}) = \frac{1}{Z}e^{\bvec{b}^{T}\bvec{h} + \bvec{c}^{T}\bvec{v} +
    \bvec{v}^{T}\bvec{W^{v}}\bvec{h} + \bvec{h}^{T}\bvec{W^{h}}\bvec{h}},
\end{displaymath}

where $\bvec{h}$ and $\bvec{v}$ are binary vectors, and $\bvec{W^{v}},
\bvec{W^{h}}$ are matrices modeling the interactions between visible units and
hidden units, hidden units and hidden units respectively. For simplicity of
notation, a feedforward network notation is used, but indices of hidden neurons
actually represent channel indices.


The competition statistic prior is endorsed by setting a fixed
between-hidden-neuron weights.
We divide hidden units into groups and set

\begin{displaymath}
  W^{h}_{ij} =
  \begin{cases}
    -\infty & \text{hidden units $i, j$ are in the same group} \\
    0 & \text{hidden units $i, j$ are in different groups}. \\
  \end{cases}
\end{displaymath}

It indicates that two hidden units in the same group cannot be activated
simultaneously (i.e. $h_i = h_j = 1$) and hidden units in different groups have
no dependency. Note that it is possible that no units is active, which is
denoted as the ground state. By setting the between-hidden-neuron weights, the
neurons with the highest activation value will be enlarged, while others
squashed. As with the traditional Boltzmann machine, temperature may be added
to intensify or temper competition. The special Boltzmann machine here is named
competitive group-restricted Boltzmann machine.


\subsection{GSMax Activation Function}
\label{sec:gsmax-activ-funct}

Given that during training, what we are interested in is the posterior
probability of hidden neurons being active, and we have purposefully kept the
inference simple, the proposed prior simplified to an activation function, and
can be put in any intermediate layers of neural network, and trained with back
propagation, which we prove in this section. We denote this new activation
function as Group Softmax (GSMax).

Denote $\bvec{h}_g$ as the vector consisting of hidden units in group $g$, and
$\bvec{h}_{-g}$ as that of hidden units outside group $g$. According to
the Boltzmann machine defined above, since $W^{h}_{ij} = 0$ when $i, j$ are not
in the same group, $\bvec{h}_g$ and $\bvec{h}_{-g}$ are independent given
visible units $\bvec{v}$. Therefore, we have

\begin{displaymath}
  p(\bvec{h} | \bvec{v}) = p(\bvec{h}_{g} | \bvec{v})p(\bvec{h}_{-g} | \bvec{v}),
\end{displaymath}
\begin{displaymath}
  p(\bvec{h}_{g} | \bvec{v}) = \frac{1}{Z_{g}}e^{\bvec{b}_{g}^{T}\bvec{h}_g  + \bvec{v}^{T}W^{v}_{g}\bvec{h}_g + \bvec{h}_g^{T}W^{h}_{g}\bvec{h}_g}.
\end{displaymath}

$\bvec{h}_{g}$ is a binary vector representing the dimensions of $\bvec{h}$
that corresponds to units in group $g$. Its length is equal to the group size
of $g$. $\bvec{b}_{g}$ is a vector composed of the elements of $\bvec{b}$ that
correspond to units in group $g$. $\bvec{W}^{v}_{g}$ is a matrix composed of
the columns $\bvec{W}^{v}$ that corresponds to units in group $g$. Similarly,
$\bvec{W}^{h}_{g}$ is submatrix of $\bvec{W}^{h}$ and its rows and columns
correspond to the units in group $g$.

Since $W^{h}_{ij} = -\infty$ when $i, j$ are in the same group,
$e^{\bvec{h}_g^{T}W^{h}_g\bvec{h}_g} = 0$ and therefore $p(\bvec{h}_{g} |
\bvec{v}) = 0$, as long as $\bvec{h}_g$ has more than one elements equal to 1,
i.e. more than one hidden units being activated. $\bvec{h}_g$ has to be one of
$\{\bvec{e}_i\}_{i=0}^{|g|}$. The dimensionality of each $\bvec{e}_i$
is $|g|$.  $\bvec{e}_0$ is an all-zero vector. $\bvec{e}_i$ has the $i$th
element equal to 1 and all others equal to 0. Denote $p(\bvec{h}_{gi} = 1 |
\bvec{v})$ as the posterior probability of hidden unit $i$ in group $g$ being active
given visible units $\bvec{v}$,  and it's also the response on neuron $i$ after GSMax. We have
\begin{displaymath}
  \fontsize{9.5pt}{10.5pt}
  p(\bvec{h}_{gi} = 1| \bvec{v})
                               = \frac{p(\bvec{h}_{g} = \bvec{e}_i|\bvec{v})}{
                                 \sum\limits_{k=0}^{|g|}p(\bvec{h}_{g} =
                                 \bvec{e}_k|\bvec{v})}
                                 = \frac{e^{b_{gi} + \bvec{v}^{T}W^{v}_{gi}}}{
                                 1 + \sum\limits_{k=1}^{|g|}e^{b_{gk} + \bvec{v}^{T}W^{v}_{gk}}}
\end{displaymath}

where $W^{v}_{gi}$ represents $i$th column of $\bvec{W}^{v}_{g}$.

In conclusion, we can see the posterior probability distribution of $\bvec{h}_g$ over
$\{\bvec{e}_i\}$ given the input $\bvec{v}$ is a Softmax function augmented
with a ground state that indicates no active hidden neurons.

\section{Related works}
\label{sec:related_works}

In this section, we review in detail existing works that use unsupervised
priors to associate complex patterns with internal neurons of neural
networks. The network used is either trained in a supervised manner, or
simplified from a supervised network for the purpose of investigation.

\cite{Mallat2012a} puts forward a simplified convolutional network with clear
mathematical interpretation in each of its components named Scattering Network
(ScatNet). ScatNet hierarchically cascades handcrafted wavelet filters
(defined on translation, rotation and scale groups), modulus non-linearity and
subsampling averaging pooling, but does not combine channels. A hidden neuron
(the activation after pooling) in ScatNet could be understood as a
representation of a concept that is invariant to actions of the group, so it
associates patterns of the input samples that are
group symmetries of each other. While without pooling, each hidden neuron associates
with a variation (a symmetry) of a certain prototypical sample. In
\cite{Oyallon2013}, they demonstrate that if limited to two layers, features
obtained by scattering transform are able to perform as well as a supervised
trained neural network on object recognition datasets. \cite{Anselmi2015}
proves that an invariant and selective (with respect to a group) signature of
an image could be obtained through pooling.
Similarly, that signature associates a hidden neuron with patterns that are
group symmetries of that sample. To investigate suitable mathematical
constructions that associate patterns in the higher layers, \cite{Bruna2013}
and \cite{Mallat2016} consider semidirect product of groups, while
\cite{Anselmi2015a} considers reproducing kernels over probability
distribution.

\cite{Gens2014} \cite{Cohen2016} \cite{Dieleman2016} apply pre-defined affine
transformation groups on feature maps (or equivalent filters), to
extend convolutional neural network beyond translation group and achieve good
results.

Complementary to the above approaches, in this work, we work from endorsing
priors on the relationship between elements of orbits to let filters associate
with known variations of concepts, then analyze the learned neurons and
filters.

\section{Experiments}
\label{sec:experiments}

In this section, we train supervised neural network with the proposed association
prior and investigate learned filters and neurons. First, we look at
filters learned in the first layer, then at the neurons of the penultimate
layer. Note that the two sections below uses different networks, the details of
which are given in the appendices.

\subsection{First layer discovers affine groups}
\label{sec:first-layer-disc}

In this section, we trained a commonly used supervised neural network on
CIFAR10 \cite{Krizhevsky2009} with the proposed prior applied to each
layer, and visualize the learned filters of the first layer to explore learned
filters. To make the competition among group elements sharply contrasted, we
choose the first layer's group size as $2$. Please refer to the appendices for
detailed network architecture and hyperparameters.

\begin{figure}[h]
  \centering
  \includegraphics[width=0.8\linewidth]{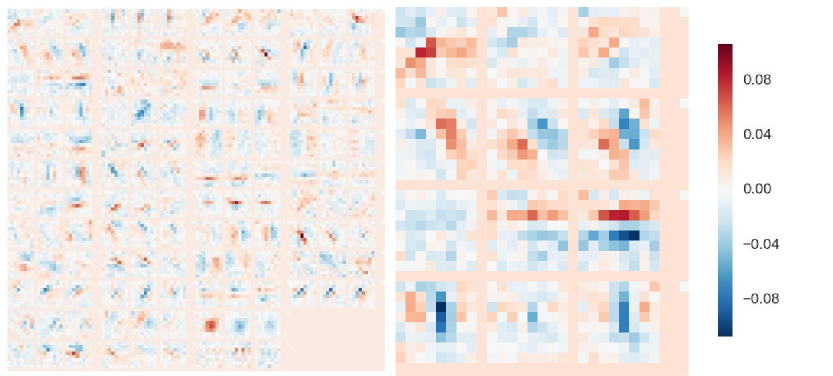}
  \caption{Left: visualize groups of filters learned in the first layer of the
network trained on CIFAR10. The group size is $2$. Each filter has three
channels. Three horizontally consecutive patches in the same row are three
2D-filters of a 3D filter. Two vertically consecutive filters in the same
column are in the same group. Right: Two groups are selected and plotted in a
larger size to better visualize the details.}
  \label{fig:filter_visual}
\end{figure}

Figure \ref{fig:filter_visual} plots several groups of filters learned in the
first layer of the network. It is observed that filters in the same group tend
to capture orthogonal visual patterns: a vertical edge tends to be in the same
group with a horizontal edge; a diagonal edge in the same group with an
anti-diagonal edge and so on. Therefore, they are complementary. The
relationship between those complementary filters possibly forms symmetry
groups. For instance, the relationship between a vertical edge and a horizontal
edge aforementioned may form a rotation group, thus a hidden neuron is able to
associate with a rotation/variation of a filter. Though previously affine
groups such as rotation group have also been observed in traditional neural
network, but we cannot anticipate or control the relationship between which
filters may form a symmetry group. While here the forming of groups conforms
with the way we divide neurons into groups. It is interesting that by endorsing
a statistics prior and purely through learning, neural networks can associate
physical variations with filters.  In this case,
the cluster of variations of an edge is the lowest level concepts a neural
network learns.

\subsection{Penultimate neurons implicitly learns sub-class concepts}
\label{sec:neur-lmpl-learns}

In this section, we investigate the filter learned in the higher layer. Due to
the composition nature of neurons in the higher layer, it is hard to deduce or
visualize learned filters corresponding to neurons. To work around, we look at
concepts that are represented by neurons, and work on a problem where the
invariant relationship between orbit elements are the membership of being in
the same super-class. In the problem, samples have a two-layer hierarchical
labels --- a super-class label and a sub-class label. We apply the proposed
prior and find the network is able to associate neurons with sub-class concepts
when trained only with the super-class labels. While without the prior
proposed, no clusters of sub-class concepts emerge.

\subsubsection{Methods}
\label{sec:methods}

In the problem, each super-class consists of sub-classes that do not overlap
with sub-classes of other super-classes, and each sample does not belong to
more than one sub-class or super-class. Dataset conforms with the above
requirements are CIFAR100 \cite{Krizhevsky2009}. We also build a similar
dataset with CIFAR100 from ImageNet \cite{Deng2009}.

We apply GSMax on the penultimate layer, and compute representation of
super-class concepts through the degenerated linear filter described in
\Cref{sec:group-assoc-prior-1}. The group division accords with the super-class
structure. For example, in CIFAR100, the penultimate layer has $100$ neurons,
the last layer has $20$ neurons, and neurons that belong to the same super class
are in the same group. Max pooling are performed by groups to build
representations of concepts of super-classes. The network is trained with only
super-class labels. The network architecture we use is wide residual network
\cite{Zagoruyko2016}. To investigate the effect of the association prior
endorsed by GSMax, we also did a control experiment. It has the same
architecture, but with no GSMax association prior applied on the penultimate
layer. To make the result easier to relate to current techniques, we also train
a Alex-Net \cite{Krizhevsky2012} style network. Please refer to the appendices
for network architecture, hyperparameters and details of the new dataset.

After finishing training the network, to associate hidden neurons with
sub-class concepts, we find the sub-class label that was mapped to a neuron
most of the times in all forward propagation, adjusted by super-class
labels. More specifically, each neuron in the penultimate layer has a list of
sub-class labels mapped to it. Initially, the list was empty. A forward
propagation was done for all test samples. For each sample, we got the
activation of the penultimate layer, and extracted neurons that correspond to
sub-classes of the super class label of this sample. Then we found the neuron
that has the largest activation value, and put the sub-class label of this
sample into the neuron's list. After all sampled had been propagated, the
sub-class label with the largest frequency in the list is taken as the true
sub-class of this neuron.

\subsubsection{Results}
\label{sec:results}

\begin{table}[h]
  \centering
  \begin{tabular}{l l l}
    \toprule
    \textbf{Architecture}     & \textbf{CIFAR-100} & \textbf{Small ImageNet} \\
    \midrule
    Maxout                    & $61.43$           & -           \\
    AlexNet                   & $50.31_{\pm0.15}$ & $47.16_{\pm0.41}$ \\
    \textbf{Implicit Learned} & $43.96_{\pm0.85}$ & $35.40_{\pm0.62}$ \\
    Control Group             & $22.47_{\pm3.26}$ & $9.83_{\pm1.76}$ \\
    Chance Level              & $20.00$           & $17.33$           \\
    \bottomrule
  \end{tabular}
  \caption{Accuracy of implicitly learned clusters of sub-classes
(in bold), and comparison between implicitly learned clusters of sub-classes
without GSMax (control group), chance level, currently supervised techniques, Alex Net and Maxout
Network. Accuracy of all networks is reported with standard deviations for
three repeated experiments, except that the result of Maxout network is
directly retrieved from the paper. '-' means the result is not available.}
  \label{tab:results_acc}
\end{table}

The results are summarized in Table \ref{tab:results_acc}. Considering that we
mask out neurons that are not in the super-class of the propagated sample when
mapping neurons to sub-classes, the chance level of classifying a sample to its
right sub-class is calculated by averaging inverse of the sub-class number in each
super class. As we could see, without any sub-class labels during training, the
neurons in the penultimate layer get a far beyond chance level accuracy. It proves a
network is able to associate a hidden neuron with a sub-class variation in a super-class concept with
only association priors. Such result is further strengthened by the fact that
without the GSMax association prior that characterizes relationship among the
sub-class neurons, the accuracy is around chance level or drastically lower
than chance level, implying no sub-class concepts emerges. The final accuracy
even may be put at the end of the performance ladder of currently pure
supervised approach, as we could see the accuracy gap between implicitly
learned clusters and Alex Net is on par with the gap between Alex Net and
Maxout network \cite{Goodfellow2013a}.

\section{Conclusion}
\label{sec:discussion}

We propose with proper priors, neural networks can associate neurons in the intermediate layers with
real-world concepts. It
implies accompanying top level labels, neural networks can discover known
concepts from data without labels, which may lead to truly unsupervised
learning. We developed a preliminary prior that fulfills this purpose. We
verified it experimentally by endorsing the proposed prior on neural networks,
and found the prior helps neural networks learn physical variations,
i.e. variations related to symmetry groups, and semantic variations,
i.e. sub-class variations in a super-class.

The proposed prior is still a crude characterization of relationship between
orbit elements to promote order during learning. An interesting future work
is to explore more general and accurate priors that may lead to adaptive
creation of concepts, when available neurons are inadequate to encode concepts
in samples.


\bibliography{../mendeley.bib/library}

\begin{thebibliography}{10}

\bibitem{LeCun2015}
Y.~LeCun, Y.~Bengio, {\em et~al.}, ``{Deep learning},'' {\em Nature}, vol.~521,
  no.~7553, pp.~436--444, 2015.

\bibitem{Krizhevsky2012}
A.~Krizhevsky, I.~Sutskever, {\em et~al.}, ``{ImageNet Classification with Deep
  Convolutional Neural Networks},'' in {\em NIPS}, 2012.

\bibitem{He2016}
K.~He, X.~Zhang, {\em et~al.}, ``{Identity Mappings in Deep Residual
  Networks},'' in {\em ECCV}, 2016.

\bibitem{Miller1972}
W.~Miller, {\em {Symmetry groups and their applications.}}
\newblock Academic Press, 1972.

\bibitem{Mallat2016}
S.~Mallat, ``{Understanding deep convolutional networks.},'' {\em Philosophical
  transactions. Series A, Mathematical, physical, and engineering sciences},
  vol.~374, no.~2065, 2016.

\bibitem{Brahma2015}
P.~P. Brahma, D.~Wu, {\em et~al.}, ``{Why Deep Learning Works: A Manifold
  Disentanglement Perspective},'' {\em IEEE Transactions on Neural Networks and
  Learning Systems}, pp.~1--12, 2015.

\bibitem{Gardner2015}
J.~R. Gardner, M.~J. Kusner, {\em et~al.}, ``{Deep Manifold Traversal: Changing
  Labels with Convolutional Features},'' pp.~1--11, 2015.

\bibitem{Montufar2014}
G.~Montufar and R.~Pascanu, ``{On the number of linear regions of deep neural
  networks},'' {\em NIPS}, 2014.

\bibitem{Zeiler2014}
M.~Zeiler and R.~Fergus, ``{Visualizing and understanding convolutional
  networks},'' {\em ECCV}, 2014.

\bibitem{Glorot2011}
X.~Glorot, A.~Bordes, {\em et~al.}, ``{Deep Sparse Rectifier Neural
  Networks},'' {\em AISTATS}, 2011.

\bibitem{Carandini2012}
M.~Carandini and D.~J. Heeger, ``{Normalization as a canonical neural
  computation.},'' {\em Nature reviews. Neuroscience}, vol.~13, no.~1, 2012.

\bibitem{Mallat2012a}
S.~Mallat, ``{Group Invariant Scattering},'' {\em Communications on Pure and
  Applied Mathematics}, vol.~LXV, pp.~1331--1398, jan 2012.

\bibitem{Oyallon2013}
E.~Oyallon, S.~Mallat, {\em et~al.}, ``{Generic Deep Networks with Wavelet
  Scattering},'' in {\em ICLR Workshop}, 2014.

\bibitem{Anselmi2015}
F.~Anselmi, J.~Z. Leibo, {\em et~al.}, ``{Unsupervised learning of invariant
  representations},'' {\em Theoretical Computer Science}, jun 2015.

\bibitem{Bruna2013}
J.~Bruna, A.~Szlam, {\em et~al.}, ``{Learning Stable Group Invariant
  Representations with Convolutional Networks},'' in {\em ICLR}, 2013.

\bibitem{Anselmi2015a}
F.~Anselmi, L.~Rosasco, {\em et~al.}, ``{On Invariance and Selectivity in
  Representation Learning},'' {\em Information and Inference, Special Issue:
  Deep Learning}, may 2016.

\bibitem{Gens2014}
R.~Gens and P.~M. Domingos, ``{Deep Symmetry Networks},'' {\em NIPS}, 2014.

\bibitem{Cohen2016}
T.~S. Cohen and M.~Welling, ``{Group Equivariant Convolutional Networks},'' in
  {\em ICML}, 2016.

\bibitem{Dieleman2016}
S.~Dieleman, J.~{De Fauw}, {\em et~al.}, ``{Exploiting Cyclic Symmetry in
  Convolutional Neural Networks},'' in {\em ICML}, 2016.

\bibitem{Krizhevsky2009}
A.~Krizhevsky, ``{Learning Multiple Layers of Features from Tiny Images},''
  tech. rep., 2009.

\bibitem{Deng2009}
J.~D.~J. Deng, W.~D.~W. Dong, {\em et~al.}, ``{ImageNet: A large-scale
  hierarchical image database},'' {\em CVPR}, 2009.

\bibitem{Zagoruyko2016}
S.~Zagoruyko and N.~Komodakis, ``{Wide Residual Networks},'' tech. rep., 2016.

\bibitem{Goodfellow2013a}
I.~J. Goodfellow, D.~Warde-Farley, {\em et~al.}, ``{Maxout Networks},'' in {\em
  ICML}, 2013.

\bibitem{Goodfellow2013b}
I.~Goodfellow and D.~Warde-Farley, ``{Pylearn2: a machine learning research
  library},'' tech. rep., 2013.

\bibitem{Abadi2015}
M.~Abadi, A.~Agarwal, {\em et~al.}, ``{TensorFlow : Large-Scale Machine
  Learning on Heterogeneous Distributed Systems},'' tech. rep., 2015.

\end{thebibliography}
\bibliographystyle{short_ieeetr}

\section{Appendices}
\label{sec:appendices}

\subsection{Experiment details for section 4.1}
\label{sec:exper-deta-sect}

Network is trained on CIFAR10 dataset. The image data are global contrast
normalized and ZCA whitened as in \cite{Goodfellow2013a}. We use exactly the
same architecture with Maxout from \cite{Goodfellow2013b}:
8C192-4MP2-8C384-4MP2-8C384-2MP2-F2500-F10 \footnote{Filter numbers for GSMax
is slightly adjusted to make group sizes divide output channel numbers}. where
8C192 means convolution layer with kernel size 8 and output channel number 192;
4MP2 means max pooling with kernel size 4 and stride 2; F10 means a fully
connected layer with output channel number 10.  Activation function is applied
after pooling layer. Dropout is applied to all the layers with keep ratio 0.5,
except for the visible unit(0.8).  Weight decay is used instead of max
norm. The base learning rate is set to 1, and decays with a rate of 0.1 every
25 epoch. Momentum optimizer with momentum 0.5 is used. GSMax group sizes are
$[2, 11, 8, 50]$, and temperature is set to 0.5.

\subsection{Experiment details for section 4.2}
\label{sec:exper-deta-sect-1}

We experimented on the CIFAR100 dataset \cite{Krizhevsky2009}. It is a dataset
of hierarchical labels that exactly conforms to the requirement discussed in
Section \ref{sec:methods}. It has 100 classes containing 600 images
each, 500 as training images and 100 as testing images. The 100 classes are
grouped into 20 super-classes. As in \cite{Goodfellow2013a}, we preprocess the
data using global contrast normalization and ZCA whitening. During training,
normally used mild data augmentation is done, aka translations and horizontal
flipping.

We also build a new dataset from ImageNet. The dataset is similar with
CIFAR100, and conforms with our requirement. It has 84321 training images and
3300 test images. The training images are from the training set of ImageNet
Large Scale Visual Recognition Challenge 2012 and testing images are from its
validation set. The small ImageNet has 66 sub-classes, and 10 super-classes
overall. During training, we used bounding boxes when available. Images are
resized to $32 \times 32$. The same data augmentation with CIFAR100 are
used. The dataset will be publicly available.

The two datasets used the same network. The network architecture is based on
wide residual network from \cite{Zagoruyko2016}, which consists of an initial
convolution layer, followed by three stages of $2n$ convolution layers using
$k_i$ filters at stage $i$, followed by a global pooling layer, an inner
product layer with GSMax as its activation function, a Maxout layer and a
Softmax layer ($6n+4$ in total). The first convolution in each stage $i > 1$
uses a stride of $2$, so the feature map sizes are $32, 16, 8$ for each three
stages. We use WRN-28-10, where $n = 4$, and $k_i = 160, 320, 640$. The first
convolution layer has $16$ filters. The neurons in the last inner product layer
are divided into groups according to the super-class structure. In the case of
CIFAR100, neurons are divided into 20 groups, each of which has 5 neurons. The
maxout layer has the same group size. Similarly for the Small ImageNet
dataset. The training procedure was reproduced as close as possible from
\cite{Zagoruyko2016} in Tensorflow \cite{Abadi2015}. Super-class labels are
applied to the last Softmax Layer.

\end{document}